\documentclass{svproc}
\usepackage{cite}
\usepackage{url}
\usepackage[detect-none,binary-units=true]{siunitx}
\DeclareSIUnit{\nothing}{\relax}
\usepackage{amsmath} 
\usepackage{graphicx}

\newcommand{\tbd}[1]{{\textcolor{black}{#1}}}

\usepackage{glossaries}
\newacronym{tof}{ToF}{Time-of-Flight}
\newacronym{ml}{ML}{Machine Learning}
\newacronym{ekf}{EKF}{extended Kalman Filter}
\newacronym{fov}{FoV}{field-of-view}
\newacronym{soc}{SoC}{System-on-Chip}
\newacronym[plural=MCUs,firstplural=microcontroller units (MCUs)]{mcu}{MCU}{microcontroller unit}
\newacronym[plural=CNNs,firstplural=convolutional neural networks (CNNs)]{cnn}{CNN}{convolutional neural network}
\newacronym{slam}{SLAM}{Simultaneous Localization and Mapping}
\newacronym{lidar}{LiDAR}{Light Detection and Ranging}
\newacronym{uart}{UART}{Universal Asynchronous Receiver/Transmitter}
\newacronym{imav}{IMAV}{International Micro Air Vehicles}
\newacronym{soa}{SoA}{State of the Art}
\newacronym{uav}{UAV}{Unmanned Aerial Vehicle}
\newacronym{ssd}{SSD}{Single-Shot multibox Detector}
\newacronym{vio}{VIO}{visual-inertial odometry}
\newacronym{qat}{QAT}{quantization aware training}
\newacronym{iot}{IoT}{Internet of Things}
\newacronym{cf}{CF}{Crazyflie}
\newacronym{OA}{OA}{Obstacle Avoidance}
\newacronym{ble}{BLE}{Bluetooth Low Energy}
\newacronym{pulp}{PULP}{Parallel-Ultra-Low-Power}
\newacronym{mcl}{MCL}{Monte Carlo Localizaiton}

\begin{document}
\mainmatter              
\title{GAP9Shield: A 150GOPS AI-capable Ultra-low Power Module for Vision and Ranging Applications on Nano-drones}

\titlerunning{GAP9Shield}  
%
\author{Hanna Müller\inst{1} \and Victor Kartsch\inst{1}
\and Luca Benini\inst{1}\inst{2}}
\authorrunning{Hanna Müller et al.} 
%
\tocauthor{Hanna Müller, Victor Kartsch, Luca Benini}
\institute{Integrated Systems Laboratory, ETH Z{\"u}rich, Z{\"u}rich, Switzerland,\\
\email{\{hanmuell, victor.kartsch, lbenini\}@iis.ee.ethz.ch}\\ 
\and
DEI, University of Bologna, Bologna, Italy\\
}

\vspace{-5mm}
\maketitle              
\vspace{-2mm}

\begin{abstract}
\vspace{-6mm}
The evolution of AI and digital signal processing technologies, combined with affordable energy-efficient processors, has propelled the development of both hardware and software for drone applications. Nano-drones, which fit into the palm of the hand, are suitable for indoor environments and safe for human interaction; however, they often fail to deliver the required performance for complex tasks due to the lack of hardware providing sufficient sensing and computing performance. Addressing this gap, we present the GAP9Shield, a nano-drone-compatible module powered by the GAP9, a 150GOPS-capable SoC. The system also includes a 5MP OV5647 camera for high-definition imaging, a WiFi-BLE NINA module, and a 5D VL53L1-based ranging subsystem, which enhances obstacle avoidance capabilities. In comparison with similarly targeted state-of-the-art systems, GAP9Shield provides a 20\% higher sample rate (RGB images) while offering a 20\% weight reduction. In this paper, we also highlight the energy efficiency and processing power capabilities of GAP9 for object detection (YOLO), localization, and mapping, which can run within a power envelope of below 100 mW and at low latency (as 17 ms for object detection), highlighting the transformative potential of GAP9 for the new generation of nano-drone applications.

\end{abstract}

\section{Introduction}
\vspace{-2mm}
The evolution of drone technology is marked by swift progress and expansion across diverse sectors, including military, industrial, and commercial, providing solutions for navigation, secure communication, reconnaissance, package delivery, smart healthcare, and precision farming. Enhanced computational capabilities are enabling the development of advanced functionalities in nano-drones, now capable of autonomously exploring and interacting with the environment while being safe humans nearby and capable of operating indoors, even in confined spaces~\cite{palossi2017self, UAV_safety_rescue, frontnet}.

Among the many available nano-drones systems, the 27g, 10cm \gls{cf} quadcopter has gained widespread adoption in the research community due to its open-source, modular architecture, which supports the quick development and testing of drone applications. 
Regarding the expansion modules, the Multi-ranger and AI decks are the most critical for deploying cutting-edge \gls{OA} and image recognition missions. The Multi-Ranger shield is a 5D VL53L1-based lightweight low-power omnidirectional range device, enabling applications to perform computationally inexpensive bio-inspired obstacle avoidance tasks efficiently\cite{lamberti2023bio}. The AI-deck is an AI vision-based engine featuring a low-power QVGA grayscale camera (Himax HM01B0) and GAP8, an 8-core \gls{mcu} (GAP8) for efficient onboard processing. 


While the AI deck exhibits sufficient processing power and sensing for missions in controlled environments, its performance is not sufficient in more complex scenarios and tasks due to its monochrome, low-resolution camera, and the limitations of the GAP8 in floating-point processing capabilities, memory size, and operational frequency. Integrating efficient processors and cameras is essential to enhance and fully exploit the capabilities of tiny drones. Similarly, considering the fundamental requirement for \gls{OA} during image acquisition and processing, combining range and high-resolution vision acquisition and processing into a single system is desirable, which could also lead to weight and size optimizations.

GAP9, the latest \gls{pulp} processor from GreenWaves Technologies, represents a significant leap forward in addressing the processing power breach for embedded systems. Furthermore, in addition to an extensive array of standard peripherals, it also features a MIPI CSI2 interface, allowing compatibility with high-definition cameras. Such features set the stage for developing a new generation of powerful AI-driven nano-drones with robust and reliable performance.

In response to the evolving demands and potential unlocked by AI, hardware, and imaging sensing innovations, this paper introduces the GAP9Shield. The GAP9Shield is a GAP9-based \gls{cf} pluggable shield incorporating a high-definition OV5647 camera and the WiFi-BLE-capable NiNA W102 module, enabling efficient onboard vision-based processing. The device can also transmit 3-channel RGB QVGA images at a frame rate 20\% higher (7FPS) than the conventional single-channel, monochrome AI-deck while also allowing streaming VGA images at \tbd{4FPS}. The shield also integrates a 5D VL53L1-based ranging subsystem for omnidirectional distance measurements, achieving a 20\% weight reduction compared to the combined AI-deck+Ranger Deck setup. Additionally, this work highlights the energy efficiency and computational capabilities of GAP9 in object detection (YOLO), localization, and mapping that can be run on GAP9 with a power envelope below 100mW and a latency ranging from 18-250ms. As a final contribution, we plan to provide the complete platform schematics and PCB diagrams as open-source resources, aiming to foster the development of next-generation nano-drone applications.

\section{System Description}
\vspace{-2mm}
\begin{figure}[t]
\centerline{\includegraphics[width=1\columnwidth]{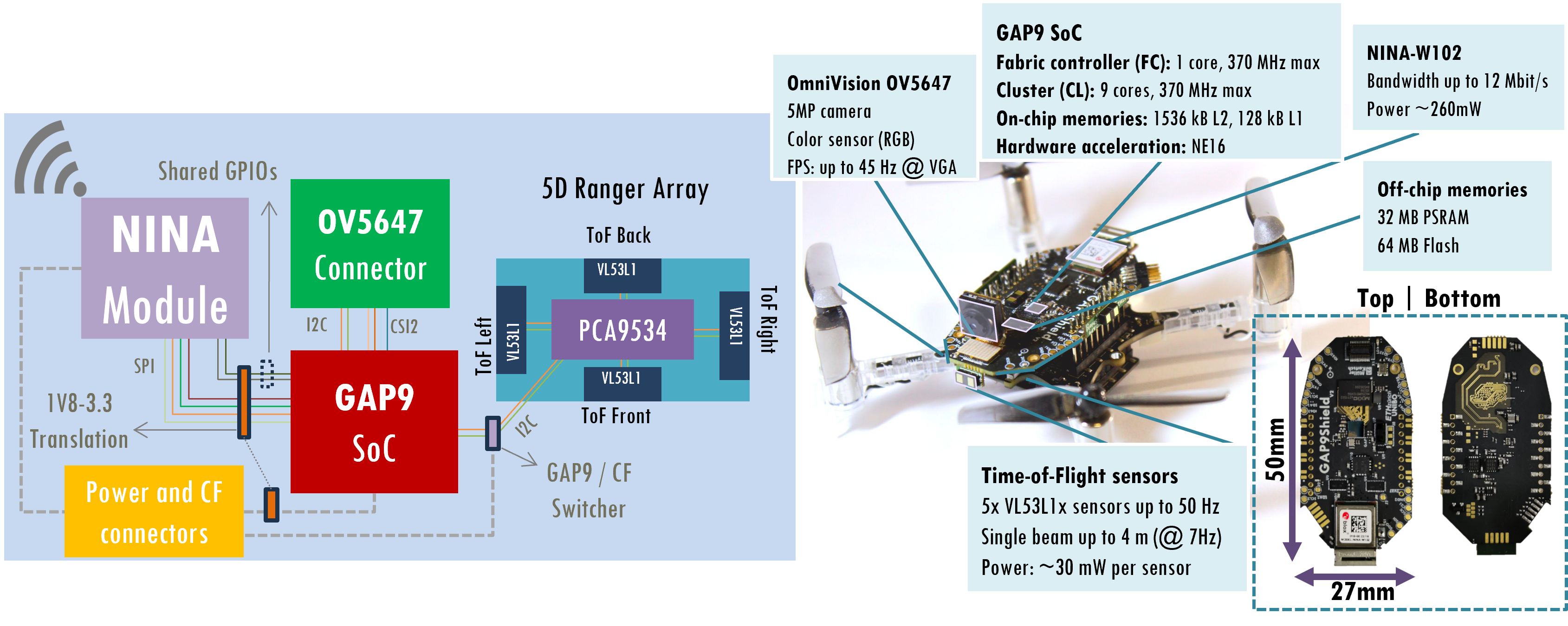}}
\caption{System overview (and block diagram) of the GAP9Shield on a \gls{cf} drone.}
\vspace{-5mm}
\label{fig:shield}
\end{figure}

Fig. \ref{fig:shield} presents a block diagram of the complete system and a summary of the features of each element within the shield. All components are integrated into a 6-layer 50x27 PCB. The following provides additional details of each component.

\subsection{GAP9 SoC} 
\vspace{-2mm}
\label{subsec:GAP9}
\begin{sloppypar}
The GAP9 \gls{soc}, by GreenWaves Technologies, features a 9-core RISC-V compute cluster with AI acceleration (NE16), transprecision floating-point support (IEEE 32-bit and 16-bit and bfloat16), which allow for 330~$\mu$W/GOP of energy efficiency and 15.6 GOPs for DSP and 32.2 GMACs for ML task of performance. Additionally, it features a single-core RISC-V controller to offload tasks to the cluster and manage peripherals. GAP9 supports various interfaces, including Serial Audio Interfaces and a CSI-2 camera interface, making it versatile for connecting to new interfaces such as current HD cameras. GAP9 features dynamic frequency (up to 370MHz), voltage scaling, automatic clock gating, and a sleep power as low as 45~$\mu$W, which allows application-specific fine-tuning of resources for optimized energy consumption. The \gls{soc} features 1.6MB of L2 RAM and 2MB of non-volatile in-package memories. On GAP9Shield, GAP9 is coupled with an APS256XXN-OBR PSRAM memory from (AP Memory) for an additional 256~Mbit of volatile memory and the MX25UM51245GXDI00 (Macronix) for a supplementary 512~Mbit of nonvolatile memory.



\end{sloppypar}

\subsection{Camera} 
\label{subsec:mems}
\vspace{-2mm}
GAP9Shield is coupled with the OV5647, a high-performance, low-voltage 5-megapixel CMOS image sensor, allowing for automatic image control while providing several image formats (QSXGA to QVGA) that can be sampled at various frame rates (15fps to 120 fps for QSXGA and QVGA, respectively). The camera supports 8/10bit raw RGB output formats that can be transferred to the host device through interfaces such as SCCB, DVP, or MIPI/CSI (single or dual lines). The OV564 is coupled to GAP9 through a single CSI2 on the GAP9Shield.

\subsection{WiFi module}
\vspace{-2mm}
\label{subsec:WiFi_module}
The NINA-W102 is a compact (10mm x 14mm), stand-alone multi-radio module integrating an ESP32 MCU with 2.4 GHz Wi-Fi, Bluetooth wireless capabilities, and a 3dBi-gain PIFA antenna. The ESP32 MCU is a dual-core SoC, and in our implementation, WiFi and SPI tasks (to communicate with GAP9) are split among the cores to maximize data throughput.

\subsection{5D Ranging Array}
\vspace{-2mm}
\label{subsec:Ranging_array}
GAP9Shield also integrates an array of 5 range sensors, pointing to the right, left, front, back, and up directions. Ranging is achieved with the VL53L1 \gls{tof} laser-ranging sensor (ST micro), allowing fast and accurate distance ranging over 400cm with full \gls{fov} at a maximum sample rate of up to 60Hz. On the GAP9Shield, the array of VL53L1s can be read directly through GAP9 or the STM32 \gls{mcu} on the \gls{cf}.


\section{Results}
\vspace{-2mm}
\subsection{System Characterization}
\vspace{-2mm}
\textbf{Physical Properties:} 
We measured our device's weight and volume at approximately 6 grams and 4050 mm3 (3x50x27 mm3), respectively. Compared to the existing \gls{cf} setup with similar functions (AI and multi-ranger decks), our device is 15\% lighter and 30\% smaller in volume\footnote{The AI and ranger decks together weigh about 7 grams and occupy 6480 mm3.}, which helps improving flight performance and battery life of the \gls{cf}.
\textbf{5D Ranging Array:} We characterized the VL53L1 sensors on the shield, employing a setup where the sensors are multiplexed over the I2C connection. The highest range frequency achieved was approximately \tbd{40SPS}, with the maximum power consumption recorded at \tbd{150mW}.
\textbf{OV5647:} We characterize QVGA and VGA resolutions for their suitability in applications with limited resources and greater image detail in more sophisticated applications, respectively. Our tests show frame rates of \tbd{15fps} for QVGA and 45fps for VGA. The lower-than-expected frame rate for QVGA stems from a driver issue, whereas VGA's performance matches the expected outcomes based on the manufacturer's specs for a single-channel CSI2. 
Power usage was recorded at \tbd{80mW} for QVGA at 15fps and \tbd{120mW} for VGA at 45fps. \textbf{WiFI Throughput and Image Streaming:}
We evaluated the NINA module's performance by sending TCP and UDP packets to a remote PC acting as a server on an existing network. The module achieved speeds of 8Mbit/s for TCP and \tbd{12Mbit/s} for UDP, with power consumption around \tbd{250mW} for TCP and \tbd{260mW} for UDP. Image streaming via TCP yielded a 3-channel RGB JPEG QVGA frame rate approximately 20\% higher than the AI-deck (around $~$7FPS) and 4FPS for the VGA resolution.




\subsection{GAP9 Performance for Nano-drone Applications}
\vspace{-2mm}
The GAP9Shield (in prototype version) has already successfully been used for many applications, such as object detection with YOLO variants, localization, and \gls{slam}. 
\textbf{Object detection:} 
In \cite{zimmerman2024icra}, a slightly modified YOLOv5 architecture was deployed on a previous version of the GAP9shield, reaching an execution time of 38ms while consuming 2.5mJ. In \cite{moosmann2023ultraefficient}, different YOLO variants were compared between three edge vision hardware platforms, one of them GAP9. The inference of their smallest network with an acceptable accuracy takes 17ms and consumes 1.59mJ.
\textbf{\gls{mcl}:} 
In \cite{zimmerman2024icra}, \gls{mcl} is used alongside YOLO-based object detection on GAP9 for localization with semantic cues onboard nano-drones. \gls{mcl} can run at sensor rate (15 Hz for the \gls{tof} and odometry update and 5 Hz for the camera update) while using an average of 23 mW during MCL execution. 
\textbf{SLAM:} 
In \cite{niculescu2023nanoslam}, the authors deploy NanoSLAM on the GAP9Shield within a power envelope of only 87.9 mW in less than 250 ms.

\section{Conclusions}
\vspace{-3mm}
This paper presented the GAP9Shield, a compact, lightweight, sensor-rich, versatile edge-AI platform for nano-drones. It can be used for dataset collection with its WIFI streamer at up to 7fps at below \tbd{260mW} as well as for onboard deployment of \gls{soa} algorithms, as YOLO, \gls{mcl} and \gls{slam} at below 100mW. 
GAP9Shield hence, unlocks new opportunities in sectors like greenhouse farming, inspection, and accurate image recognition and navigation in cluttered conditions.

\bibliographystyle{IEEEtran} 
\bibliography{bib}

\end{document}